\documentclass[sigconf,article]{acmart}
\renewcommand\footnotetextcopyrightpermission[1]{} 
\usepackage{tikz}
\usetikzlibrary{arrows.meta,positioning,fit,backgrounds}
\usepackage{graphicx}

\begin{document}

\title{SciTrue: Reliable Scientific Claim Validation with Frontier and Open Language Models at the NTCIR SciClaimEval Task}

\author{Qiming Bao}
\authornote{Equal contribution; co-first authors.}
\affiliation{\institution{University of Auckland}\country{New Zealand}}
\email{qiming.bao@auckland.ac.nz}

\author{Neşet Özkan TAN}
\authornotemark[1]
\affiliation{\institution{University of Auckland}\country{New Zealand}}
\email{neset.tan@auckland.ac.nz}

\author{Siyuan Wang}
\affiliation{\institution{The Chinese University of Hong Kong}\country{Hong Kong}}
\email{siyuanwang1997@gmail.com}


\author{Mark Gahegan}
\affiliation{\institution{University of Auckland}\country{New Zealand}}
\email{m.gahegan@auckland.ac.nz}

\begin{abstract}
We describe the SciTrue team's participation in both subtasks of the NTCIR-19
SciClaimEval task~\cite{sciclaimeval}, which asks systems to verify scientific claims
against the tables and figures of a paper. Rather than tuning a single model, we
benchmark eleven frontier and open multimodal models under one honest, per-sample
protocol and combine them with light, transparent post-processing. On the official,
blind test leaderboard (Section~\ref{sec:results}), SciTrue placed first by a clear
margin in three of the four evidence-category/subtask combinations, and tied for first
on the primary metric in the fourth. Three findings explain the result. First, strong
instruction-tuned models are already competitive: Claude Opus~4.8 and Gemma-4-31B each
exceed the strongest public baseline (o4-mini), and GPT-5.5 and Claude Fable~5 lead
both subtasks (97.7 on Subtask~2). Second, the task's pairing structure is the largest
lever: a \emph{leak-free pair prior} that recovers the Supported/Refuted pairing from
the claim text alone (a visible field) and assigns Supported to the higher-confidence
evidence raises Subtask-1 pair-accuracy from 72.2 to 93.5, far more than any model swap
or ensemble weighting. Third, a case-by-case audit finds that most residual errors are
visually-undetectable label-mapping swaps or dataset label noise, so measured accuracy
understates the true ability and the fixable-by-modeling headroom is small. Controlled
fine-tuning, distillation, and agentic consistency-checking support the same
conclusions, and we document throughout a measurement leak---label information reaching
a system through the packaging of the data rather than its content---in which the
released file ordering encodes the label, including one instance that briefly misled our own pipeline.
\end{abstract}


\maketitle
\pagestyle{plain} 

\section*{Team Name} 
SciTrue

\section*{Subtasks}
Subtask 1 (Claim Label Prediction) and Subtask 2 (Claim Evidence Prediction).

\section{Introduction}
The quantitative backbone of a scientific paper lives in its tables and figures, and a
claim's credibility often hinges on whether it faithfully reflects that evidence.  As
submission volume and AI-assisted writing grow, automatically checking claims against
the underlying evidence is an increasingly practical need, a multimodal counterpart to
textual and tabular fact-checking~\cite{fever,scifact,tabfact}. The NTCIR-19 SciClaimEval
task~\cite{sciclaimeval} evaluates exactly this: Subtask~1 labels requires participants to label each claim as \emph{Supported}/\emph{Refuted} given one table or figure; Subtask~2 requires participants to correctly identify which of two evidence images supports the claim. Team \textit{SciTrue} entered both, and we omit further task
background, as it is described in detail by the overview paper~\cite{sciclaimeval}.
Our guiding observation is that, with today's instruction-tuned vision-language models,
the bottleneck is no longer raw image reading but \emph{how predictions are combined and
how the task's structure can be teased out and used}. This splits into three coupled questions: (1) which models to trust, (2) how to fuse them, (3) how to exploit the Supported/Refuted pairing, studied under one honest protocol and leading to five contributions:
\begin{itemize}
  \item a uniform, \textbf{honest per-sample benchmark} of eleven frontier and open
  multimodal models, on which Opus~4.8 and Gemma-4-31B both surpass the o4-mini
  baseline;
  \item a \textbf{leak-free pair prior} that recovers claim pairs from text and turns
  two absolute judgments into one relative ranking, lifting pair-accuracy to 93.5;
  \item a documented \textbf{measurement-leak correction}: position-based tie-breaks
  inflate dev scores but do not transfer to the pair-hidden test set, so we report
  only leak-free numbers;
  \item a \textbf{fine-tuning study} and \textbf{failure audit} showing most residual
  errors are visually-undetectable label swaps or dataset noise, not a perception gap,
  leaving little headroom for better modeling; and
  \item an \textbf{agentic consistency checker} that cross-references the evidence
  image against the claim's source paper to catch tampering the ensemble misses, and a
  \textbf{distillation study} showing this behavior partially transfers to small open
  VLMs via QLoRA.
\end{itemize}
\section{Related Work}

\paragraph{Claim verification and fact-checking.}
Automated fact-checking is studied extensively in the textual setting, from
open-domain verification over Wikipedia~\cite{fever} to scientific claims against
abstracts~\cite{scifact} and structured tables~\cite{tabfact}. A parallel line of work
asks where the claims themselves come from and how a verifier should justify its
verdict: claims can be generated automatically from multi-choice
questions~\cite{multi2claim} or extracted directly from the full text of scientific
papers~\cite{echoes}, and verification can be framed so that the reasoning chain, not
only the label, is faithful to the retrieved evidence~\cite{faithful}. Our team name
comes from SciTrue~\cite{scitrue-demo}, a deployed system for evidence-grounded claim
verification in science, and this body of work is surveyed and extended
in~\cite{tan-thesis}.  The SciClaimEval challenge extends the setting beyond the text of the paper to include figures and tables, i.e. a multimodal regime and related to recent efforts to benchmark
VLMs as fact-checkers~\cite{mfcbench}. Like table fact verification~\cite{tabfact}, it
is contrastive: each claim pairs with original and minimally tampered evidence, which
we exploit directly.

\paragraph{Chart/table understanding and vision-language models.} Reasoning over a
figure means reading axes, legends, and panels and comparing trends, probed by chart/plot
QA~\cite{chartqa} and document/table understanding~\cite{docvqa} benchmarks. A
common approach first converts the image evidence to a structured form~\cite{deplot,tapas} then
reasons over text; we instead pass the official structured table data as an auxiliary
channel and rely on native visual reasoning for figures. The VLMs we draw on span
instruction-tuned~\cite{llava} architectures through today's open and frontier
families: Gemma~\cite{gemma}, Qwen-VL~\cite{qwen2vl,qwen3vl,qwen35}, GLM-V~\cite{glmv},
InternVL~\cite{internvl}, Llama-3.2-Vision~\cite{llama3}, GPT-4~\cite{gpt4},
Claude~\cite{claude}, and o4-mini~\cite{openai_o4mini}. We benchmark eleven models drawn
from these families under one protocol rather than committing to a single model.

\paragraph{Reliability, fusion, and benchmark quality.} Demanding benchmarks from VQA~\cite{vqa}
to expert-level~\cite{mmmu} multimodal reasoning have driven
progress, supported by fragility research based on grounding failures~\cite{pope} and on LLMs
proving weaker abstract reasoners than their fluency suggests~\cite{gendron2024llm};
the SciClaimEval challenge adds further focus, since a model that faithfully reads a
tampered-but-internally-consistent image is correct about the image yet wrong about the
claim. Our post-processing draws on
reasoning-plus-acting~\cite{react}, multi-sample/model aggregation~\cite{distillation},
and low-rank adaptation~\cite{lora,qlora,cora}. Our contribution is less a new technique than
showing that a structure-aware prior dominates all of them here. Our failure audit also
echoes work on pervasive test-set label errors~\cite{labelerrors}: many of our
``failures'' are mislabeled or visually-undetectable manipulations, not model weakness.

\section{Task and Data}
SciClaimEval draws claims and evidence from research papers across three domains:
machine learning, natural language processing, and biomedicine (PeerJ). Each evidence
item is a table or a figure, provided as a PNG; for tables the organizers also release
the underlying structured data (LaTeX/HTML/JSON). In \textbf{Subtask~1} (Claim Label
Prediction) a system labels a claim as Supported or Refuted given one evidence item;
the development set has 747 claims and the test set 917. Crucially, the data are
\emph{contrastive}: a claim is paired with its original evidence (Supported) and a
minimally tampered copy (Refuted), so the development set forms 352 such pairs (plus a
few supported-only singletons). The primary metric is \textbf{pair-accuracy}: a pair
counts as correct only if both members are labeled correctly, which rewards systems
that genuinely discriminate the tampering rather than guessing a marginal. In
\textbf{Subtask~2} (Claim Evidence Prediction) a system is shown two evidence images
and selects the one that supports the claim (352 development / 436 test samples,
scored by accuracy). We abbreviate the development set as \emph{dev} and
pair-accuracy as \emph{pair-acc} throughout. The test labels and the pairing field are withheld; results are returned by
the challenge organizers only after the competition has ended.

\section{Methods}

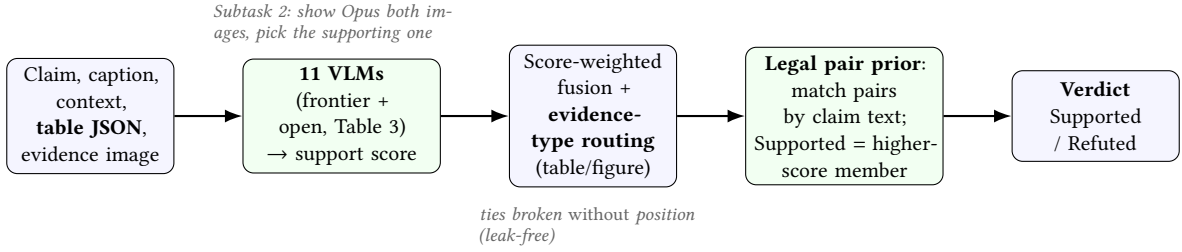
\begin{figure*}[t]
\centering
\begin{tikzpicture}[
  font=\small, >=Latex, node distance=7mm and 9mm,
  box/.style={draw, rounded corners, align=center, minimum height=12mm,
              text width=20mm, inner sep=3pt, fill=blue!4},
  big/.style={draw, rounded corners, align=center, minimum height=12mm,
              text width=24mm, inner sep=3pt, fill=green!5},
  lab/.style={font=\footnotesize\itshape, text=black!60}]
\node[box] (in) {Claim, caption, context, \textbf{table JSON}, evidence image};
\node[big, right=of in] (models) {\textbf{11 VLMs}\\ (frontier + open, Table~\ref{tab:leaderboard})\\ $\to$ support score};
\node[box, right=of models] (route) {Score-weighted fusion + \textbf{evidence-type routing} (table/figure)};
\node[big, right=of route] (pair) {\textbf{Legal pair prior}: match pairs by claim text; Supported $=$ higher-score member};
\node[box, right=of pair] (out) {\textbf{Verdict}\\ Supported / Refuted};
\draw[->,thick] (in)--(models);
\draw[->,thick] (models)--(route);
\draw[->,thick] (route)--(pair);
\draw[->,thick] (pair)--(out);
\node[lab, below=1.5mm of route, text width=30mm] {ties broken \emph{without} position (leak-free)};
\node[lab, above=1mm of models, text width=34mm] {Subtask~2: show Opus both images, pick the supporting one};
\end{tikzpicture}
\caption{The SciTrue Subtask-1 pipeline: eleven models each provide a support score;
we fuse the strongest, route by evidence type, and apply a leak-free pair prior
turning two absolute judgments into one relative ranking. Subtask~2 is handled
directly by Opus~4.8.}
\label{fig:pipeline}
\end{figure*}

Figure~\ref{fig:pipeline} summarises our system. We deliberately avoid task-specific
training in the main pipeline and instead compose strong off-the-shelf models with
three transparent operations: score-level fusion, evidence-type routing, and a pair
prior derived from the dataset's construction. Each operation reads only ffelds released for both splits: the ensemble's support scores, the evidence type
 ffag evi\_type (for routing), and the claim text (for pair matching). Each is published for the test split as well as the
development split. None of the three operations touches \texttt{claim\_id\_pair},
which the organizers withhold at test time, or the order in which rows appear in the
development file. The gains are therefore reproducible on the blind test set rather
than artefacts of the development release.
We describe each component below, followed by the controlled fine-tuning study and the evaluation protocol used to assess accuracy.

\subsection{Models and inference}
We evaluate multimodal models spanning frontier and open systems. The frontier tier is
Anthropic's Claude Opus~4.8~\cite{claude} (queried through its CLI, requiring no local
GPU), the OpenAI o4-mini~\cite{openai_o4mini} baseline reported by the organizers, and
two models we evaluated once they became available, later incorporated into
additional submitted runs (Section~\ref{sec:fable5}): OpenAI's GPT-5.5~\cite{gpt5} and
Anthropic's Claude Fable~5~\cite{fable5}, the latter also queried through its CLI.
The open tier covers four families run locally on A100-80GB GPUs: Google's
Gemma-4~\cite{gemma} (31B and the E4B variant); the Qwen vision-language line, namely
Qwen3-VL~\cite{qwen3vl} (8B and the 30B-A3B mixture-of-experts), the newer
Qwen3.5/Qwen3.6 multimodal models~\cite{qwen35} (9B, 35B-A3B, and the latest
35B-A3B), and Qwen2.5-VL~\cite{qwen2vl} (used for the fine-tuning study); and Zhipu's
GLM-4.6V-Flash~\cite{glmv}. For completeness the organizers' baselines also include
InternVL3.5~\cite{internvl} and Llama-3.2-Vision~\cite{llama3}. Every model receives
the claim, the caption, the surrounding context when the dataset marks it necessary,
and the evidence image; for tables we additionally supply the official structured table
data as an auxiliary text channel. Each model is prompted to return a JSON verdict with
a support score in $[0,1]$, which we use both as the label and as the confidence weight
for fusion. We recover the verdict from the last balanced JSON object in the output,
repairing quote and backslash errors, and fall back to thresholding the score at $0.5$
when no object parses.

\paragraph{Implementation.} Every model receives the same system prompt. It casts the
model as a scientific fact-checker, directs it to read axis labels, legends,
sub-figure tags and exact cell values, to attend to comparison words (\emph{highest},
\emph{outperforms}, \emph{increases}) and to metric direction, and to reply with only
a JSON object carrying a brief reasoning field, a support score in $[0,1]$, and a
label. No few-shot examples are given and the prompt is never tuned per model, so
differences across the eleven systems reflect the models rather than the prompting.
We disable the extended reasoning trace on the Qwen3.5 and Qwen3.6 models, which is
three to four times faster and made a full development sweep feasible on shared
hardware; all other models run in their default configuration. Images are resized where needed to keep memory within budget on shared GPUs (1536px longest side
for the Transformers-based models, Qwen's native min/max-pixel budget for the Qwen line, 1568px
for Opus; GPT-5.5/GPT-5.6 Sol are sent unresized); large MoE
models are sharded across two GPUs, with 8-bit quantisation where a model would
otherwise not fit. Opus~4.8 runs through its CLI, reading images from disk, needing
no local accelerator. Across the eleven models the score-threshold fallback ffres on well under one output in twenty. All models see identical inputs, and every raw response is
stored, so fusion, routing, and the pair prior are computed afterwards from those
stored responses rather than by re-querying the models. Reproducing our results
requires neither GPUs nor API access.

\subsection{Honest per-sample evaluation}
A subtle pitfall shapes how we report every number. The development file lists each
claim pair with the Supported member first (claim ids are assigned sequentially, S then
R). Consequently, any post-processing that resolves a pair by \emph{row position}---a
tie-break that defaults to the first member, say, or always presenting the first member
as ``image~1'' to a comparison model---is not reading the evidence at all. It is
recovering the gold label, the ground-truth answer distributed with the development
set, from the row order alone. Such a rule scores well on the development set while
measuring nothing: the test set withholds \texttt{claim\_id\_pair} and does not
guarantee the same ordering, so the advantage disappears the moment it is scored. We therefore adopt a strict rule: all reported numbers are raw per-sample
predictions, and wherever the pair prior must break a tie it does so \emph{without}
using position. The ordering is a leak in the standard sense: it carries label
information through the packaging of the data rather than its content, so a system can
score well without performing the task. We quantify its size in
Section~\ref{sec:results}, and we
keep the one method that benefits from ordering (pairwise-gated fusion) as a separate,
clearly-labeled submission rather than our headline result.

\subsection{Ensembling and evidence-type routing}
We fuse models at the score level by summing their (sign-oriented) support scores.
Curating the three strongest models clearly beats pooling all of them, as the weaker
models inject noise rather than useful diversity. Moreover, almost every model we tested
is markedly weaker on figures than on tables (the newest frontier model is the
exception), and the per-type rankings differ. We exploit this with \emph{evidence-type
routing}: table evidence is scored by Opus~4.8, Gemma-4-31B, GPT-5.5 and Fable~5, and
figure evidence by those four plus GLM-4.6V-Flash, which is competitive on figures
alone. Because \texttt{evi\_type} is released for the test split as well as the
development split, the same rule applies unchanged to the blind test set.

\subsection{Legal pair prior}
By construction, most claims appear twice: once with the original evidence (labeled
Supported) and once with a tampered version of the same evidence (labeled Refuted).
The organizers hide the pairing field (\texttt{claim\_id\_pair}) in the test set, but
the two members share \emph{identical claim text}. Grouping the development claims by
exact claim string therefore recovers the pairing exactly---352 of 352 two-member
groups, each with one Supported and one Refuted member, identical to the hidden
field---while using only a visible input. Formally, for a recovered pair $(a,b)$ with
ensemble support scores $s_a,s_b$, we predict $\hat{y}_a=\text{Supported}$,
$\hat{y}_b=\text{Refuted}$ if $s_a>s_b$, and the reverse otherwise. This replaces two
\emph{absolute} judgments (``does this
evidence support the claim?'') with one \emph{relative} judgment (``which of the two
better supports it?''), which the models answer far more reliably---especially for
under-specified claims that are hard to adjudicate in isolation. The $\sim$43 unpaired,
supported-only claims (texts that appear once) are labeled Supported. We stress that
the prior is permitted under the task rules: it uses the claim text (a visible
field) and the publicly
documented one-Supported-one-Refuted construction, never the hidden pairing.

\subsection{Pairwise comparison and gating (and a position-bias caveat)}
For pairs where the ensemble is inconclusive, we also tried a direct Subtask-2-style
comparison: show Opus both sets of evidence and ask which supports the claim.  This is used only when
the ensemble margin is small (gated fusion). On dev this appeared to reach 96.9
pair-acc, but a swapped-order probe revealed a $\sim$15-point bias toward ``image~1''
(97\% in the original order vs.\ 82\% when the Supported member is image~2). Assuming the same construction convention holds on test as on dev (\S4.2)---unverifiable, since test carries no labels---this inflation most plausibly reflects ordering, not verification skill; we therefore submit gated fusion only as a separate
run (Run~1.5) and keep the leak-free pair prior as our primary result. An
order-balanced version (ask both orders, commit only if they agree) is the proper
de-biased form.

\subsection{Image-versus-paper consistency check}
\label{sec:harness}
Since most residual errors (\S\ref{sec:fail}) are self-consistent label-mapping swaps
with no in-image signal, and each claim's source paper is provided
(\texttt{paper\_path}), we built an agentic checker that retrieves the paper excerpt
most lexically/numerically overlapping with the claim/caption (the four best-scoring
paragraphs, truncated to 700 characters each). Learned attention offers an alternative
to this lexical heuristic~\cite{tan-attention}; we did not pursue it, since the relevant
passage already fell inside that window. The checker then asks Claude Opus~4.8 (via CLI, with file-system Read access
to the image) whether the image and excerpt are \emph{consistent}, reframing
verification as \textbf{tamper detection} rather than the ``does this evidence support
the claim'' framing that weakened our negative-result retrieval experiment
(\S\ref{sec:results}). Used alone, it is a noisy classifier (it both recovers failures
and disturbs some already-correct pairs), so we add its score as a
\emph{supplementary} signal on top of the routed ensemble rather than replacing it.

\subsection{Distilling the consistency check}
\label{sec:distill}
To test whether this ``read image, cross-check paper, flag contradiction'' behavior can
be taught to an open model rather than requiring Claude at inference, we collect the
checker's structured traces (what the image shows, what the paper says, consistency,
verdict) on the LoRA train fold, keep only traces whose verdict matches gold, and
QLoRA-fine-tune two open VLMs---Qwen3.5-9B and Gemma-4-31B~\cite{gemma}---to reproduce
them from the same inputs (claim, caption, retrieved excerpt, evidence image). We
evaluate on a held-out fold (72 pairs, disjoint from the LoRA fold) against the same
zero-shot model and the Claude teacher, scoring uncovered/tied pairs at half credit
(unlike the strict pair-accuracy used elsewhere).

\subsection{LoRA fine-tuning (dev cross-validation)}
SciClaimEval has no training split, so to study fine-tuning honestly we partition the
development set 80/20 (train/held-out), keeping both pair members in the same fold to
avoid leakage (596 train, 151 held-out). We QLoRA-fine-tune Qwen2.5-VL-7B~\cite{qwen2vl}
(4-bit NF4, LoRA adapters of rank 16, $\alpha=32$, dropout 0.05 on the attention/MLP
projections, training only the assistant label token, two epochs on one A100-80GB),
reading the ``Supported''-vs-``Refuted'' next-token probability as the support score,
and evaluate on the
held-out fold against the identical model used zero-shot (the same adapter, disabled).

\section{Experiments and Results}
\label{sec:results}
\paragraph{Setup.} We evaluate on the full development set (747 Subtask-1 claims, 352
Subtask-2 pairs) with the official metrics; Subtask-1's primary metric is
\textbf{pair-accuracy} (both members of a pair correct), reported alongside macro-F1.
Open models run locally on A100-80GB GPUs through Hugging Face Transformers, while
Opus~4.8 is queried through its CLI without a local GPU. All scores are raw per-sample
predictions; we apply no pair-forcing that would key on the file ordering (Sec.~4.2).
As a sanity check, our run of Qwen3-VL-8B reproduces the official baseline, confirming
the harness. \textbf{Official test results.} The organizers found our original
Subtask-1 table-type prompts were not pure PNG (they included injected structured-data
text), disqualifying them from a PNG-only category (Subtask~2 always was PNG-only). We
resubmitted five true PNG-only Subtask-1 runs (\S\ref{sec:runs-png}), scored on
2026-08-02: Claude Fable~5 alone reached 98.01 macro-F1 /
98.15 pair-accuracy, the best of the five, with the 5-model ensemble close behind
(97.79/97.92). Fable~5 and GPT-5.6 Sol (each alone) transfer cleanly from their
same-protocol dev numbers (95.3/95.7 and 95.3/96.0 respectively), confirming a small
dev$\to$test gap once pair-forcing protocol is held fixed.

\paragraph{Final official leaderboard.} The organizers published the full leaderboard
at \url{https://sciclaimeval.github.io} on 2026-08-02: three evidence-input categories
for Subtask~1 (PNG, JSON, TeX/HTML, reflecting the resubmission split above) and one for
Subtask~2 (Table~\ref{tab:finalboard}). SciTrue placed first, by a clear margin, in
three of the four: Subtask-1 JSON (98.4 pair-acc vs.\ 93.2 runner-up), Subtask-1
TeX/HTML (98.4 vs.\ 97.7), and Subtask-2 PNG (98.4 vs.\ 98.2). In the fourth,
Subtask-1 PNG, our top run ties Bonn-Juelich Informatics at 98.2 on the primary
pair-accuracy metric (no tie-break rule is published, so we report a tie rather than
claim a win) while leading all four secondary metrics (98.0 vs.\ 97.2). A 3-of-4 clear
win, tied on the primary metric in the fourth, is strong blind-test validation of this
paper's dev-set findings.

\begin{table}[t]\centering\small
\caption{Final official leaderboard (organizers, 2026-08-02). Primary metric:
pair-accuracy (Subtask~1), accuracy (Subtask~2). $\ddagger$ = tie, no published
tie-break rule.}
\label{tab:finalboard}
\begin{tabular}{llccl}
\toprule
Subtask & Category & SciTrue & Runner-up & Result\\
\midrule
1 & JSON & 98.4 & 93.2 (Black Socks) & \textbf{1st}\\
1 & TeX/HTML & 98.4 & 97.7 (Bonn-Juelich) & \textbf{1st}\\
1 & PNG & 98.2 & 98.2 (Bonn-Juelich) & \textbf{tied 1st}$\ddagger$\\
2 & PNG & 98.4 & 98.2 (Bonn-Juelich) & \textbf{1st}\\
\bottomrule
\end{tabular}
\end{table}

\subsection{Submitted runs}
We submitted several runs per subtask (the form permits multiple, and a new run does
not overwrite earlier ones); each is named by its method so the leaderboard can
distinguish them. Table~\ref{tab:runs} lists every submitted run with its dev score.
Our original primary (reportable) runs were the leak-free pair prior for Subtask~1 and
Opus-4.8 for Subtask~2. We also submitted a position-leveraged variant (Run~1.5) as a
separate, clearly-labeled run: it exploits the claim-id ordering artifact rather than
verification skill and is not our headline result. As GPT-5.5 and Claude Fable~5
became available to us, we submitted additional runs incorporating each
(Runs~1.6--1.7, 2.3--2.4), discussed in Section~\ref{sec:fable5}.

\label{sec:runs-png}
\paragraph{Post-deadline PNG-only extension (Runs 1.8--1.12, Subtask~1 only).} Under a
short organizer-granted extension (\S\ref{sec:results}) we reran Subtask~1 with the
structured-data channel removed for table-type claims and a newly available model,
GPT-5.6 Sol~\cite{gpt5}, and submitted five true PNG-only runs (Table~\ref{tab:runs},
rows 1.8--1.12; official test scores discussed in \S\ref{sec:results}).

\begin{table*}[t]\centering\small
\caption{All submitted runs (dev; pair-forced, matching what each file submits). Pair
= Subtask-1 pair-accuracy; Acc = Subtask-2 accuracy. $\dagger$ = position-leveraged
(not headline). Runs 1.8--1.12 are the post-deadline PNG-only resubmissions
(\S\ref{sec:runs-png}).}
\label{tab:runs}
\begin{tabular}{p{0.72\textwidth}c l}
\toprule
Run / method & Subtask & Dev\\
\midrule
1.1--1.4 Opus+Gemma+Qwen3.5-35B ensemble + routing + legal pair prior + singletons (leak-free; full progression in Table~\ref{tab:ablation}) & 1 & 93.5 pair (F1 95.2)\\
1.5 + pairwise-gated fusion $\dagger$ & 1 & 96.9 pair\\
1.6 + GPT-5.5 (leak-free) & 1 & 95.6 pair\\
1.7 + Claude Fable~5 (leak-free) \textbf{[best main-track]} & 1 & \textbf{96.2} pair\\
1.8 Claude Fable~5 alone, PNG-only & 1 & 95.7 pair\\
1.9--1.11 + GPT-5.5 / + Fable~5 / + GPT-5.6 Sol ensembles, PNG-only (leak-free) & 1 & --- \\
1.12 GPT-5.6 Sol alone, PNG-only \textbf{[latest]} & 1 & 96.0 pair\\
2.1 Opus-4.8 & 2 & 92.6 acc\\
2.2 Opus+Gemma-4-31B+GLM-4.6V & 2 & 92.9 acc\\
2.3 GPT-5.5 alone & 2 & 97.4 acc\\
2.4 Claude Fable~5 alone \textbf{[latest]} & 2 & \textbf{97.7} acc\\
\bottomrule
\end{tabular}
\end{table*}

\begin{table}[t]\centering\small
\caption{Single-model dev results (honest per-sample). ``ours'' = run by us; numbers
refreshed 2026-08-02 after fixing two harness bugs (\S\ref{sec:results}), moving a few
models by 1--5 points. $^{*}$GPT-5.6 Sol, added later for Subtask-1 only
(\S\ref{sec:runs-png}), is not among the eleven both-subtasks models.}
\label{tab:leaderboard}
\begin{tabular}{llccc}
\toprule
Model & Src & T1 F1 & T1 Pair & T2 Acc\\
\midrule
Claude Fable~5 & ours & \textbf{86.3} & \textbf{74.4} & \textbf{97.7}\\
GPT-5.6 Sol$^{*}$ & ours & 84.5 & 70.2 & ---\\
GPT-5.5 & ours & 84.5 & 69.9 & 97.4\\
Gemma-4-31B-it & ours & 84.5 & 69.9 & 88.9\\
Claude Opus 4.8 & ours & 82.9 & 67.6 & 92.9\\
o4-mini & official & 82.9 & 68.2 & 85.2\\
Qwen3.6-35B-A3B & ours & 82.5 & 66.2 & 92.0\\
GLM-4.6V-Flash & ours & 79.6 & 60.8 & 83.2\\
Qwen3.5-35B-A3B & ours & 79.5 & 61.1 & 86.1\\
Qwen3-VL-30B-A3B & official & 76.0 & 54.8 & 54.3\\
Gemma-4-E4B-it & ours & 75.6 & 52.6 & 68.8\\
Qwen3.5-9B & ours & 73.0 & 48.3 & 81.8\\
Qwen3-VL-8B & official & 72.1 & 46.9 & 56.2\\
Qwen3-VL-8B & ours & 71.5 & 45.2 & 63.9\\
Qwen3-VL-30B-A3B & ours & 71.2 & 46.0 & 66.5\\
InternVL3.5-38B & official & 67.8 & 40.1 & 54.5\\
Llama-3.2-11B-Vision & official & 48.6 & 10.8 & 34.7\\
\bottomrule
\end{tabular}
\end{table}

\paragraph{Single models (Table~\ref{tab:leaderboard}).} Opus 4.8 and Gemma-4-31B both
exceed o4-mini; our Qwen3-VL-8B (71.5/45.2) reproduces its official baseline
(72.1/46.9), validating the harness. The newest open model we tested,
Qwen3.6-35B-A3B, is the second-strongest open system (82.5 F1 / 66.2 pair-acc), ahead
of Qwen3.5-35B-A3B and surpassing o4-mini on Subtask~2 (92.0).

\paragraph{The newest frontier models.} OpenAI's GPT-5.x line~\cite{gpt5} and
Anthropic's Claude Fable~5~\cite{fable5} became available after our original runs
(Runs 1.1--1.5, 2.1--2.2 in Table~\ref{tab:runs}); we evaluated both on the full
development set and submitted additional runs incorporating each (Runs 1.6--1.7,
2.3--2.4). \textbf{GPT-5.5} reaches 84.5 F1 / 69.9 pair-acc on Subtask~1 and 97.4 on
Subtask~2 (vs.\ 92.6 for Opus, our original Subtask-2 run), with a table-vs-figure gap
of only $+2.5$ (Table~\ref{tab:evtype}), far below every model in our original runs
($+5$ to $+14$). Swapping it into the routed ensemble lifts the leak-free pair prior
from 93.5 to 95.6 (Run~1.6).

\label{sec:fable5}
\textbf{Claude Fable~5} is stronger still (86.3 F1 / 74.4 pair-acc on Subtask~1, 97.7 on
Subtask~2, Table~\ref{tab:leaderboard}) and has the smallest table-vs-figure gap of any
model we tested ($+1.8$). Reaching this number required catching our own mistake: an
initial pass, computed from our pipeline's own prediction file, read an implausible
99.4 pair-acc; that file turned out to use the hidden \texttt{claim\_id\_pair} field for
pair-forcing and to silently default failed API calls to a low-confidence label, rather
than reflecting genuinely per-sample, per-model judgments. A model-specific usage cap
also required an automated retry loop across several hours to reach full coverage. The
corrected, fully-covered number is strong but in line with GPT-5.5 rather than an
outlier. We view this episode as a useful, if uncomfortable, illustration of this
paper's broader warning (Section~\ref{sec:results}): even a solution built around
honest, leak-free evaluation can be misled by its own tooling, and every number in this
paper that could not be cross-checked this way deserves the same scrutiny. Adding
Fable~5 to the GPT-5.5-augmented routed ensemble raises the leak-free pair prior from
95.6 to \textbf{96.2} on all 352 pairs, a smaller marginal gain than GPT-5.5's own,
consistent with diminishing returns as the ensemble's members grow individually
stronger and more correlated.

\begin{table}[t]\centering\small
\caption{Subtask-1 accuracy by evidence type, for all models we ran (refreshed
2026-08-02; see Table~\ref{tab:leaderboard}). Every model is weaker on figures, the gap
widens for smaller models, and rankings shift---motivating evidence-type routing.}
\label{tab:evtype}
\begin{tabular}{lccc}
\toprule
Model & Table & Figure & Gap\\
\midrule
Gemma-4-31B & 87.3 & 79.6 & $+7.7$\\
Claude Fable~5 & 87.1 & 85.3 & $\mathbf{+1.8}$\\
Claude Opus 4.8 & 86.3 & 77.4 & $+8.9$\\
GPT-5.5 & 85.5 & 83.0 & $+2.5$\\
Qwen3.6-35B-A3B & 84.9 & 78.1 & $+6.8$\\
Qwen3.5-35B-A3B & 82.4 & 74.3 & $+8.1$\\
GLM-4.6V-Flash & 81.5 & 76.2 & $+5.3$\\
Gemma-4-E4B & 79.7 & 68.3 & $+11.4$\\
Qwen3.5-9B & 76.6 & 66.8 & $+9.8$\\
Qwen3-VL-30B-A3B & 76.3 & 61.9 & $+14.4$\\
Qwen3-VL-8B & 75.3 & 64.5 & $+10.8$\\
\bottomrule
\end{tabular}
\end{table}

\paragraph{Tables vs.\ figures (Table~\ref{tab:evtype}).} Every model is weaker on
figures than on tables, by 1.8 to 14.4 points. The gap is by far the smallest for the
two newest frontier models (Fable~5 $+1.8$, GPT-5.5 $+2.5$) and largest for the smaller
open ones (Qwen3-VL-30B $+14.4$, Gemma-4-E4B $+11.4$), indicating that figure
understanding is exactly where capability is still increasing---and where the latest
models pull ahead. The per-type ranking also reshuffles---for
example, GLM-4.6V-Flash reads figures nearly as well as much larger models despite a
lower overall rank. Because the manipulation types that fool models (legend and category
swaps; Section~\ref{sec:fail}) occur in figures, this is also where most of our residual
errors concentrate, and it is exactly the signal evidence-type routing exploits.

\begin{table}[t]\centering\small
\caption{Subtask-1 ablation (dev). Fusion and routing add a point or two; the legal
pair prior is decisive. ``All-model'' fusion is worse than the curated trio; the last
row adds GPT-5.5 (Run~1.6).}
\label{tab:ablation}
\begin{tabular}{lcc}
\toprule
Method & Macro-F1 & Pair-Acc\\
\midrule
Best single model at the time (Gemma-4-31B) & 84.5 & 69.9\\
3-model score fusion & 85.4 & 72.2\\
\quad + evidence-type routing & 85.9 & 73.0\\
All-model fusion & 82.9 & 66.2\\
\quad\textbf{+ legal pair prior} & \textbf{95.2} & \textbf{93.5}\\
\quad\quad + GPT-5.5 (Run~1.6) & 95.7 & \textbf{95.6}\\
\bottomrule
\end{tabular}
\end{table}

\begin{center}\small
\captionof{table}{Where the pair prior's gain comes from (dev, 3-model ensemble).
Same-label = both members got the same absolute label, unresolvable without
relative comparison.}
\label{tab:pairbreak}
\begin{tabular}{lcc}
\toprule
Outcome & Pairs & \%\\
\midrule
Already correct (honest) & 253 & 71.9\%\\
Fixed by relative comparison & 80 & 22.7\%\\
Still wrong after correction & 19 & 5.4\%\\
\bottomrule
\end{tabular}
\end{center}

\paragraph{Ensembling and the pair prior (Table~\ref{tab:ablation}).} Score-weighted
fusion of the three strongest models reaches 72.2 pair-acc, and evidence-type routing
73.0; pooling all 8 models available at the time of these runs is \emph{worse} (66.2),
as the weaker systems add noise
rather than diversity. The choice of fusion rule matters little---majority vote,
score-weighting, reliability-weighting, and an exhaustive subset search all land at
72.2, and only evidence-type routing improves on them. The legal pair prior is the decisive step, lifting pair-accuracy
to \textbf{93.5} (leak-free; the leaky position-tie-break variant reads 94.9). The gain
holds across domains (pair-acc 95.8 ML, 94.5 NLP, 89.4 PeerJ), with the biomedical split being the most challenging, consistent with its denser tables and figures. Subtask~2 is already a pairwise
selection task; a three-model fusion edges out Opus alone by a small margin
(93.5 vs.\ 92.9, both refreshed post-\S\ref{sec:results}).

\paragraph{Why relative comparison works (Table~\ref{tab:pairbreak}).} The gain is
concentrated: 71.9\% of pairs are already correct under honest prediction; in
27.3\%, both members get the \emph{same} absolute label (unresolvable by any
threshold), and correcting only these recovers 94.6\% pair-accuracy---indistinguishable
from correcting all 352. A global threshold sweep instead gains at most 0.5 points
(71.9$\to$72.4, oracle-tuned), ruling out simple recalibration. Errors here are
biased---models default to ``Supported'' twice as often as ``Refuted'' when confused
(65 vs.\ 31)---and the 19 pairs relative comparison cannot fix concentrate in the
manipulation types already flagged as visually undetectable (\S\ref{sec:fail}):
legend/category swaps fail three times more often than cell-value edits, mostly in
figures.

\paragraph{Subtask 2.} The two subtasks reward different recipes. Subtask~2 already
presents both evidence images and asks which supports the claim---structurally the same
relative judgment that our pair prior reconstructs for Subtask~1. Here Opus~4.8 is
strong on its own (92.9 accuracy, vs.\ 88.9 for the best open model and 85.2 for the
o4-mini baseline); a three-model fusion improves on it by a small margin (93.5), a gain
that postdates our original submission. We therefore submitted Opus directly for
Subtask~2 and reserved fusion for Subtask~1, where the models are more complementary. The contrast also
explains why our Subtask-1 pair prior works: it imports exactly the relative-comparison
formulation that makes Subtask~2 tractable.

\paragraph{Fine-tuning.} On the held-out fold, QLoRA fine-tuning lifts Qwen2.5-VL-7B
from 77.8 to \textbf{84.7} pair-acc (+6.9 over zero-shot), a substantial gain for a few
hundred training pairs and a single A100. Yet the tuned 7B still trails the
training-free ensemble+prior (93.1 on the same pairs) by a wide margin. Fine-tuning is
a viable, cheap route to a competent standalone model, but still does not provide the best results---the
headroom lies in scale, error-diverse fusion, and the pairing structure, consistent with
our broader finding that the residual difficulty is structural, not a matter of model
adaptation.

\paragraph{Image-versus-paper consistency check.} Adding the checker's score
(Section~\ref{sec:harness}) to the routed ensemble covers 184 of 352 pairs (the
remainder have no paper excerpt with enough lexical/numeric overlap to retrieve) and
lifts pair-accuracy from 93.5 to \textbf{93.6}, or from 95.6 to \textbf{95.9} once
GPT-5.5 is included. The gain is modest in aggregate because most residual errors are
the visually-undetectable swaps identified in Section~\ref{sec:fail}, where the source
paper's prose often does not reveal the true label-to-curve mapping either; but on
failures where the paper text does state the relevant number, the checker recovers them
by catching an explicit image/paper contradiction (e.g., a table cell reads 80.52\% in
the image while the paper states 85.52\%). This turns our earlier negative result
(retrieved context \emph{degraded accuracy} when used to answer the claim directly)
into a positive
one, once the same text is used to check consistency instead.

\paragraph{Distilling the consistency check.} On the held-out fold
(Section~\ref{sec:distill}), distillation raises Qwen3.5-9B from 51.4 to
\textbf{62.5} pair-acc ($+11.1$) and Gemma-4-31B from 49.3 to \textbf{55.6} ($+6.3$),
showing that some of the ``cross-check the paper'' behavior transfers to a small open
model via a modest number of QLoRA examples. Both students still trail the Claude
teacher (77.1 on the same pairs, 84.1 on the 41/72 it actually covers) by a wide
margin, so this is a partial distillation rather than a drop-in replacement; closing
the gap is left to future work.

\paragraph{Measurement leak and de-biasing.} The position bias identified in \S4.5
also inflates the pair prior itself: with position-based tie-breaking, it reads 94.9
pair-acc, essentially the same $\sim$15-point leak as the gated-fusion probe.
Removing position from the tie-break drops this to \textbf{93.5}, which we report as
our headline; the gated variant remains a separate, labeled run (\S4.5).

\paragraph{Negative results.} Two ``more information'' fixes \emph{reduced accuracy}: feeding
retrieved paper context to the context-dependent pairs (92.2$\rightarrow$84.3) and a
legend-mapping prompt that asks the model to transcribe the legend before judging. Both
fail for the same reason---the manipulation is inside the evidence, while
external/explicit text describes the un-tampered original and thereby \emph{masks} the
tampering. This is the empirical counterpart of the audit in Section~\ref{sec:fail}:
the hard cases are self-consistent edits with no in-image signal, so adding context
cannot help and often actively misleads.

\section{Failure Analysis}
\label{sec:fail}
To understand where the remaining headroom lies, we audited the Subtask-1 failures
case by case, comparing each claim against \emph{both} the original and the tampered
image.\footnote{Audited at an earlier $\sim$92.0\% pipeline stage ($\sim$28 failures);
category proportions are assumed, not re-confirmed, at the current 93.5\% headline
(23 pairs).} They fall into three groups. (i)~\textbf{Visually-undetectable
label-mapping swaps} (most of the errors): a legend or category axis is relabeled to
produce an internally consistent chart, so the manipulation is a \emph{valid} refutation
that nonetheless leaves no in-image trace---the gold label is correct, but no
image-only model can detect it. (ii)~\textbf{Dataset label noise}: cases where the
tampering does not actually falsify the claim, so the ``Refuted'' label is questionable;
we confirmed several by hand (e.g.\ one pair whose only edit makes a $p$-value
\emph{more} significant while leaving the effect direction unchanged, so the claim still
holds) and reported the three clearest cases to the organizers, who confirmed the issue
and indicated the affected samples would be removed from the development set via a
public announcement; every dev number in this paper (747/352) predates that
removal.\footnote{Organizer correspondence, 2026-07-06/07.} (iii)~A
smaller number of \textbf{genuine model misses} on visible edits the
model should have caught (e.g.\ a row swap that does flip a ``best method'' comparison).
Two consequences follow. First, measured accuracy understates true ability, because some
``errors'' are mislabeled and others are undetectable by construction. Second, the
fixable-by-better-modeling fraction is small: most of the residual is
\emph{visually-undetectable manipulation or benchmark noise} rather than a perception
gap. This also explains our negative results: when the alteration is internal
and self-consistent, supplying more text (a retrieved paragraph, a legend-reading
prompt) only reinforces the tampered evidence rather than exposing it.

\begin{table}[t]\centering\small
\caption{Pair-accuracy by tampering type. The model handles value/structure edits
well, but \emph{label-mapping} swaps (legends, category axes) are the weak
spot---visually self-consistent.}
\label{tab:manip}
\begin{tabular}{lcc}
\toprule
Manipulation type & Pair-acc & $n$\\
\midrule
Graph swap & 100.0 & 10\\
Change cell values & 96.7 & 183\\
Graph flip & 92.5 & 20\\
Swap rows / columns & 92.5 & 53\\
Legend swap & 88.9 & 63\\
Category swap & 80.4 & 23\\
\bottomrule
\end{tabular}
\end{table}

\begin{figure}[t]\centering
\includegraphics[width=0.25\textwidth]{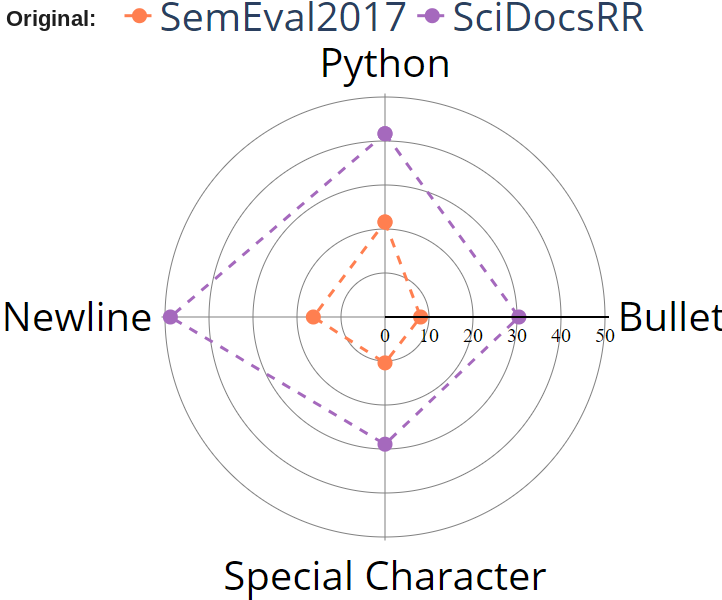}\\[2mm]
\includegraphics[width=0.26\textwidth]{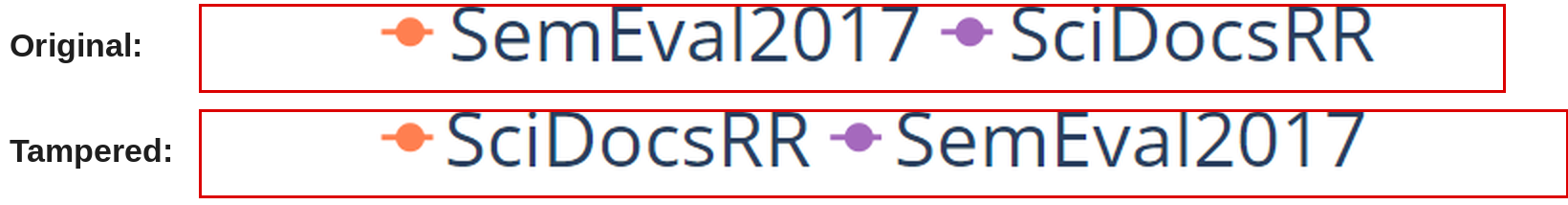}
\caption{An \emph{invisible legend swap} (pair \texttt{val\_fig\_0154/0155}, a radar
chart of formatting robustness). \emph{Top:} the original chart (gold
\emph{Supported}); the geometrically identical tampered copy is omitted.
\emph{Bottom:} the legend zoom---orange is \textbf{SemEval2017} in the original but
\textbf{SciDocsRR} in the tampered copy (red boxes), with the \texttt{Python}/
\texttt{Bullet} axis labels likewise swapped. The points never move, so the relabeling
leaves \emph{no geometric trace}, and all three ensemble models assign both members
the same label---a genuine ceiling, not a perception error.}
\label{fig:legendswap}
\end{figure}

Breaking the errors down by \emph{manipulation type} (Table~\ref{tab:manip}) sharpens
the picture: the model is very strong at catching swapped or flipped graphs and altered
cell values (92--100\%), but degrades exactly where the edit rewires the label-to-value
mapping (legend/category swaps, 80--89\%). A
swapped legend or category axis produces a chart that is internally coherent---if the
green curve is relabeled ``Mistral,'' the green curve genuinely \emph{is} Mistral in
that image---so there is no in-image signal of tampering. This is also why figures
trail tables: legends/category axes exist only in figures, which carry a traceless
tamper mode, whereas a tampered table cell is just a wrong value checkable against
the structured data we supply. We conclude that the
remaining errors are not a perception problem solvable by higher resolution or
better prompting, but would require an external consistency check (e.g.\ comparing the
rendered figure against the structured data), which we leave to future work.

A second case (\texttt{val\_fig\_0093/0094}) swaps the \textbf{GAT}/\textbf{APPNP}
legend entries across just 0.3\% of pixels, otherwise pixel-identical---the same
undetectable pattern. Visible edits like graph-flip and value-change
(Table~\ref{tab:manip}, 92--100\%) are usually recoverable: a biomedical pair
(\texttt{val\_fig\_0288/0289}) shifts two boxed groups from $\approx$100 to
$\approx$350--680, which the ensemble still misranks---a perception slip, not the
legend-swap blind spot.

\section{Discussion and Limitations}
\paragraph{Structure beats scale, honestly measured.} Exploiting the task's pairing
structure (+21 points, leak-free per \S\ref{sec:results}) dwarfs every model swap or
fine-tuning effort, but only once position-based leakage is removed---a hazard for any
task with order-encoded labels. Our audit (\S6)
finds most ``errors'' are visually-undetectable or mislabeled, not genuine misses;
organizers confirmed and removed the cases we reported, corroborating a dataset
artifact. A cleaner consistency check would re-render the chart from structured
data for a direct pixel/value comparison~\cite{deplot}; our checker's
verify-then-decide logic could itself be a learned training-time
prior~\cite{baoconflict}, and the pairwise comparison's position bias also calls
for an order-balanced fix (\S\ref{sec:results}). Our pipeline is cheap and
transparent (no local accelerator for the frontier model, one A100-80GB per open
model, no training). Limitations: the pair prior assumes the
one-Supported-one-Refuted construction holds on test (unverifiable without
labels), and several open models ran non-thinking, understating them slightly.

\section{Conclusions}
SciTrue's entry to NTCIR-19 SciClaimEval combines a frontier-plus-open ensemble,
evidence-type routing, and a leak-free pair prior (recovered from the claim text alone)
to reach 93.5 pair-accuracy on the development set. Adding GPT-5.5 and Claude Fable~5
(the strongest single models on both subtasks) raises this to 96.2, after catching a
pair-forcing bug in our own pipeline (\S\ref{sec:fable5}). These dev-set findings held
up on
the blind official test: SciTrue placed first, by a clear margin, in three of four
categories, tying for first in the fourth (Subtask-1 PNG, 98.2). A controlled
fine-tuning study and failure audit confirm most residual errors
are a benchmark ceiling, not model weakness; an agentic checker recovers some,
partially distilling into small open VLMs. We also document and correct a measurement leak in which the released
ordering encodes the label, and hope these findings are useful to future
participants.\footnote{Code and per-model predictions:
\url{https://github.com/taneset/Sci-claim}}

\bibliographystyle{ACM-Reference-Format}
\bibliography{ntcirsample}

\end{document}